\documentclass[12pt, final]{l4dc2023}
\pdfoutput=1

\usepackage[capitalize]{cleveref}
\usepackage{mathrsfs}
\usepackage{physics}
\usepackage[group-separator={,},per-mode=symbol]{siunitx}
\usepackage{algorithm,algorithmic}
\usepackage{booktabs}
\usepackage[font=small]{caption}
\usepackage{dsfont}
\usepackage{wrapfig}
\usepackage{standalone}
\usepackage{multirow}

\graphicspath{{./}{./figures/}}
\input{figures/preamble}

\newcommand{\ph}[1]{{\textbf{#1}:}} %

\title[Model-based Validation as Probabilistic Inference]{Model-based Validation as Probabilistic Inference}
\usepackage{times}

\author{\Name{Harrison Delecki} \Email{hdelecki@stanford.edu}\\
\Name{Anthony Corso} \Email{acorso@stanford.edu}\\
\Name{Mykel J. Kochenderfer} \Email{mykel@stanford.edu}\\
\addr Department of Aeronautics and Astronautics\\
Stanford University, Stanford, CA 94305, USA}

\begin{document}

\maketitle

\begin{abstract}%
Estimating the distribution over failures is a key step in validating autonomous systems. Existing approaches focus on finding failures for a small range of initial conditions or make restrictive assumptions about the properties of the system under test. We frame estimating the distribution over failure trajectories for sequential systems as Bayesian inference. Our model-based approach represents the distribution over failure trajectories using rollouts of system dynamics and computes trajectory gradients using automatic differentiation. Our approach is demonstrated in an inverted pendulum control system, an autonomous vehicle driving scenario, and a partially observable lunar lander. Sampling is performed using an off-the-shelf implementation of Hamiltonian Monte Carlo with multiple chains to capture multimodality and gradient smoothing for safe trajectories. In all experiments, we observed improvements in sample efficiency and parameter space coverage compared to black-box baseline approaches. This work is open sourced.\footnote[1]{\href{https://github.com/sisl/ModelBasedValidationInference}{https://github.com/sisl/ModelBasedValidationInference}}
\end{abstract}


\section{Introduction}\label{sec:intro}
Greater levels of automation are being considered in safety-critical applications such as self-driving cars \citep{badue2021selfdrivingsurvey} and air transportation systems \citep{straubinger2020overviewUAM}. These systems typically operate sequentially by taking actions after observing the environment. Sequential decision-making systems in safety-critical domains require thorough validation for acceptance and safe deployment. Information about potential failure modes is critical for engineers and policymakers to determine whether certain failure modes require additional engineering attention and to estimate the risk associated with deployment \citep{Corso2021survey}. Therefore, a key step of safety validation is to determine the distribution over failure trajectories, or search for possible failure modes and determine their associated likelihood.

Searching for failures in safety-critical systems is challenging for several reasons. First,  failures tend to be rare. Safety-critical autonomy is designed to be relatively safe from the outset, so naive methods like direct Monte Carlo sampling require an enormous number of samples to discover rare failures. Second, the search space is high dimensional. Autonomous systems typically have large state spaces and operate over long time horizon trajectories. Third, sequential systems can exhibit multimodal failures. Accurately estimating multimodal distributions requires methods that can explore the parameter space without converging to individual modes.

Several previous approaches to safety validation perform falsification \citep{dreossi2015efficient, Akazaki2018falsification, tuncali2019rapidly}, which aims to find a single failure example or failure mode. The main drawback of these methods is that they tend to converge to a single failure trajectory, such as the most extreme or most likely example \citep{lee2020adaptive}. The mode-seeking behavior of falsification methods is insufficient to capture the distribution over failure trajectories. Importance sampling \citep{kim2016improvingAircraftCollisionCEM, zhao2017acceleratedIS, okelly2018scalableAVTestingCEM} and Markov Chain Monte Carlo \citep{norden2019efficient, sinha2020neural} approaches have also been used to sample failure trajectories. These approaches aim to approximate the distribution over failures using samples. However, existing approaches are only applicable in low-dimensional spaces and rely on ad-hoc algorithmic modifications. These limitations make existing approaches less useful for the validation of sequential systems in a general setting.  

\begin{figure}[tbp]
    \subfigure[\small Model-based sampling of the distribution over failures.]{\includegraphics[width=0.55\textwidth]{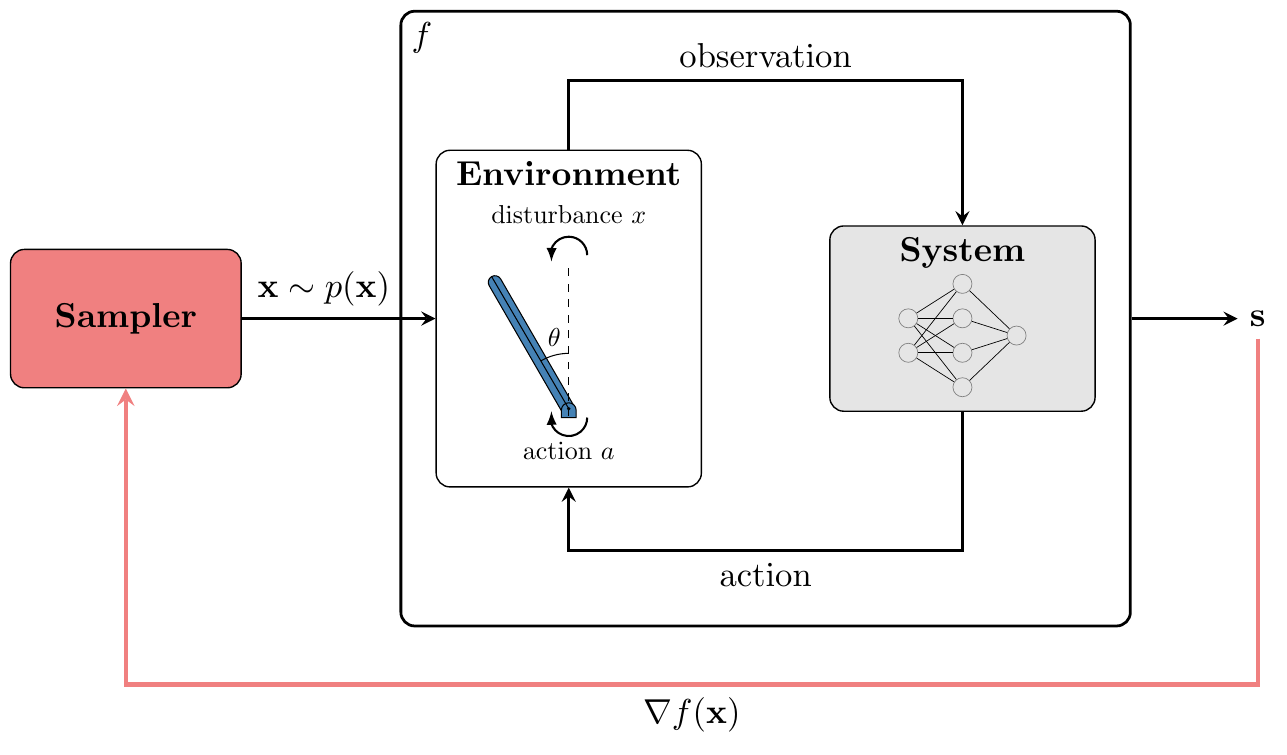}}
    \hfill
    \subfigure[\small Sampled failure trajectories of the inverted pendulum.]{\raisebox{0.2\height}{\includestandalone[width=0.35\textwidth]{pendulum_failures_hmc}}}
    \caption{A diagram of the proposed approach for estimating the distribution over failures. A sampler uses gradients of system dynamics to find disturbances $x$ that lead to a failure, such as toppling an inverted pendulum. The distribution over failures is estimated by drawing many samples that result in failure.}
    \label{fig:main}
    \vspace{-3mm}
\end{figure}

In this work, we aim to enable efficient approximation of the distribution over failure trajectories in sequential systems. First, we frame the problem as Bayesian inference of the distribution over failures due to disturbances in the environment. We then propose a modeling framework to represent the failure distribution in a probabilistic programming paradigm. This model-based representation enables us to easily use advanced inference algorithms for high-dimensional sampling problems, such as Hamiltonian Monte Carlo \citep{brooks2011handbookMCMC, hoffman2014noUTurn}. The proposed approach is illustrated in \cref{fig:main}. Our contributions are as follows:
\begin{itemize}
    \itemsep0em 
    \vspace{-1mm}
    \item We frame sampling the distribution over failures in sequential systems as Bayesian inference.
    \item We propose a model-based approach to sample from the distribution over failures.
    \item We demonstrate the proposed approach on three sample validation problems.
\end{itemize}

\section{Related Work}\label{sec:relatedwork}
Relevant related work can be grouped in three categories: white-box validation methods, falsification approaches, and sampling-based approaches.

\ph{White-box methods} Traditional methods in system validation often focus on white-box techniques, which assume that a full mathematical model of the system is known~\citep{katoen2016probabilisticModelChecking, clarke2018handbookmodelchecking}. These methods typically do not scale to large problems because they consider every possible execution of the system and the environment \citep{alur2015principlescps}. The goal of this work is similar in that we also use knowledge internal to the system to search for multiple possible failure events.

\ph{Falsification} Several recently proposed safety validation methods for sequential systems take a falsification approach. These methods use optimization \citep{deshmukh2017testing}, trajectory planning \citep{dreossi2015efficient, tuncali2019rapidly}, or reinforcement learning \citep{Akazaki2018falsification} to find disturbances in the environment that lead to any failure or the most severe failure. Adaptive Stress Testing is a reinforcement learning method that aims to find the most-likely failures according to a prescribed probability model \citep{lee2015adaptive, koren2018adaptive}. Falsification methods tend to converge to a single failure mode rather than exploring the distribution over failures.

\ph{Sampling-based approaches} Importance sampling has been applied to estimate the distribution over failures for aircraft collision avoidance systems and autonomous vehicles \citep{kim2016improvingAircraftCollisionCEM, okelly2018scalableAVTestingCEM}. The cross-entropy method (CEM) aims to iteratively learn the optimal importance sampling distribution. CEM relies on the specification of a parameterized family of distributions and may fail when the true underlying distribution is multimodal or high dimensional \citep{geyer2019cross}.  Markov Chain Monte Carlo (MCMC) has also been applied to estimate the distribution over failures \citep{botev2008efficientMultiLevelSplitting, norden2019efficient}. MCMC methods used in previous work explore the distribution using a random walk or hand-crafted proposal distributions \citep{brooks2011handbookMCMC}. While this is effective for lower dimensional problems, this exploration strategy struggles to scale to high dimensional distributions. In this work, we use Hamiltonian Monte Carlo to estimate the distribution over failure trajectories. HMC is an MCMC algorithm that uses gradients rather than a random-walk or a hand-crafted proposal to explore the distribution.

\section{Approach}\label{sec:method}
In this section, we present our framework for model-based Bayesian inference of the failure distribution in sequential systems. First, we introduce the necessary notation and definitions. Then, we describe our approach to estimate the distribution over failures through probabilistic inference.

\subsection{Formulation}
We build upon the notation of \citet{Corso2021survey} to describe the safety validation problem. Consider a system under test that takes actions in an environment after receiving observations. We define a safety property $\psi$ that we aim to evaluate for the system under test. The property $\psi$ is defined over the state trajectories of the environment, $\vb{s}=[s_1,\dots,s_t]$, where $s_t$ is the state of the environment at time $t$. If the trajectory results in a failure, we say that $\vb{s} \notin \psi$.

We perturb the environment with disturbances with the goal of inducing the system under test to violate the safety specification. A disturbance trajectory $\vb{x}$ has probability density $p(\vb{x})$. We assume that disturbances determine all sources of stochasticity in the environment. For example, sensor noise and stochastic dynamics are potential disturbances that could lead to failure.

We denote the simulation of the system under test with environment disturbances by a dynamics function $f$. Note that $f$ represents the dynamics of both the system under test and the environment. The resulting state trajectory under disturbances is written as $\vb{s} = f(\vb{x})$. Since all stochasticity in the simulation is determined by $\vb{x}$, the disturbance model $p(\vb{x})$ induces a distribution over state trajectories, $p(\vb{s})$. We seek the distribution over state trajectories that violate the safety property, the conditional distribution $p(\vb{s} \mid \vb{s} \notin \psi)$.

\subsection{Approach Overview}
Environment dynamics or characteristics of the system under test often make it impossible to express the failure distribution analytically. However, it is usually easy to sample from these systems by simulating the system under test with disturbances. Bayesian inference methods such as MCMC approximate a target distribution by drawing samples. In this work, we frame the safety validation problem as Bayesian inference of the distribution over failures.

Sampling-based Bayesian inference requires a way to draw samples from a target distribution. We represent the state trajectory distribution as a probabilistic model using probabilistic programming \citep{gordon2014}. Under this paradigm, we can describe the distribution over state trajectories as a function of random variables that represent the disturbance model. Modern probabilistic programming languages, such as Turing \citep{ge2018turing}, let us wrap existing simulations in a probabilistic program with little modification of the underlying simulation.


This probabilistic model represents the distribution over state trajectories induced by the disturbance model. We represent the distribution over failures by conditioning the probabilistic model on the desired outcome of the simulation --- that the simulated trajectory is a failure. Any safe trajectory has zero likelihood under the conditional distribution. We represent this condition using the Dirac delta distribution, whose value is zero for all state trajectories except when $\vb{s} \notin \psi$. Under this formulation, the log-likelihood of a state trajectory $\ell(\vb{s} \mid \vb{s} \notin \psi)$ is
\begin{equation}
    \ell(\vb{s} \mid \vb{s} \notin \psi) =  \log \delta(\vb{s} \notin \psi) + \log p(\vb{s})
    \label{eq:true-distribution}
\end{equation}

The main challenges associated with sampling from $p(\vb{s} \mid \vb{s} \notin \psi)$ are the distribution's high dimensionality, discontinuity, and multimodality. The following sections show how we address each of these challenges.

\subsection{Model-based Approach with Automatic Differentiation}
State trajectories in sequential systems of interest tend to be high dimensional. It is well established that black-box, gradient-free sampling from high-dimensional distributions becomes difficult with increasing dimensionality~\citep{brooks2011handbookMCMC}. Gradient-based sampling algorithms such as Hamiltonian Monte Carlo (HMC) \citep{duane1987hybrid, girolami2011riemann} use information about the shape of the underlying distribution to take large steps in parameter space, allowing them to scale to higher dimensions.

To address the challenge of sampling high-dimensional failure trajectories in sequential systems, we take a model-based approach. Gradients of the dynamics function $f$ with respect to disturbances are computed using automatic differentiation. The simulation environment and system under test are modeled in a framework compatible with automatic differentiation (AD). AD enables algorithms to compute exact gradients of the posterior likelihood density with respect to the input disturbances. We use Zygote \citep{innes2019zygote}, which performs source-to-source AD of Julia code.

In contrast with some previous approaches that assume the system is Markov \citep{corso2020ScalableDP}, our approach can handle non-Markov systems. We estimate the distribution over failures using rollouts of the system dynamics under disturbances, which fully characterize the system behavior and allows us to validate policies that depend on state estimates rather than true states.

\subsection{Smoothing Approximation for Sharp Gradients}
Gradients of the posterior density will be undefined in regions where the state trajectory is safe due to the Dirac delta in \cref{eq:true-distribution}. Additionally, the boundary between safe and failing state trajectories is discontinuous. As a result, gradient-based sampling algorithms have no information about how to improve a proposal. We relax the discontinuity by approximating the Dirac delta distribution as a one-dimensional Gaussian with a small variance $\epsilon$. We also assume that we can calculate a distance to failure $\Delta(\vb{s})$ for any safe trajectory. The likelihood of the distance to failure under the Gaussian approximation of the discontinuity is added to the likelihood of the sampled trajectory. The log-likelihood of a trajectory $\ell(\vb{s})$ under this smoothing approximation is
\begin{equation*}
    \ell(\vb{s} \mid \vb{s} \notin \psi) =  \log \mathcal{N}(\Delta(\vb{s}) \mid 0, \epsilon) + \log p(\vb{s})
\end{equation*}
The role of AD is to calculate $\nabla_{\vb{x}} \ell (\vb{s} \mid \vb{s} \notin \psi)$, which is now defined for safe and failure trajectories. The Gaussian smooths the discontinuity between the safe and failure regions of the state trajectory space while changing the density in the failure region by a constant. This smoothing approximation along with AD enables us to use gradient-based sampling methods.

\begin{wrapfigure}{r!}{0.33\textwidth}
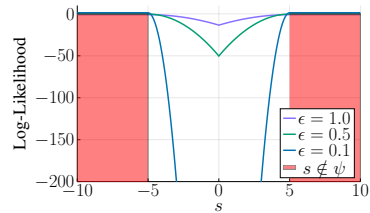

  \begin{center}
    \includestandalone[width=\linewidth, trim=0 0 0 0, clip]{figures/epsilonillustrated}
  \end{center}
  \vspace{-3mm}
  \caption{Illustration of smoothing hyperparameter $\epsilon$.}
  \label{fig:epsilon-illustrated}
\end{wrapfigure}

The variance $\epsilon$ acts as a hyperparameter, with smaller values better approximating the discontinuity and larger values yielding smoother posterior densities. Consider a validation problem for illustrative purposes only, where the state distribution is the unit Gaussian and failure occurs when $|s| > 5$. \cref{fig:epsilon-illustrated} shows the log-likelihood of single-step state trajectories for various values of $\epsilon$. As $\epsilon$ decreases towards zero, the resulting distribution is a better approximation of the true failure distribution, shown in red. Larger values of $\epsilon$ smooth the gap between the separated failure modes on the far left and right tails of the distribution. Empirically, we find values in the range $\epsilon \in [0.01, 0.1]$ work well for the problems investigated.

\subsection{Sampling for Multimodality}
The generality of the approach makes available a wide variety of off-the-shelf Bayesian inference methods. We use HMC with No-U-Turn Sampling (NUTS) \citep{hoffman2014noUTurn}, which is a general HMC variant that excels at sampling from complex, high-dimensional posterior distributions.

Multimodality in the distribution over failure trajectories poses a challenge for many MCMC algorithms~\citep{rudoy2006monte}. Sampling algorithms may converge to a single mode rather than mixing across the modes in parameter space. In our illustrative example, a single MCMC chain may only sample from the left or right mode of the failure distribution. One way to explore a multimodal space is to run many chains of MCMC from different starting points. In our sampling approach, we use multiple NUTS chains with different starting disturbance trajectories. The number of chains selected depends on the difficulty of the problem, with higher dimension and more modes generally requiring more chains.

\section{Experiments}\label{sec:experiments}

This section presents the experimental evaluation of the proposed approach. First, we discuss our evaluation metrics and baseline sampling approaches. Then, we describe the three simulation environments and policies used in validation experiments. 

\ph{Metrics and Baselines}
We evaluate the proposed approach based on failure rate, log-likelihood of failures discovered, and computation time. The failure rate is the proportion of drawn samples that result in failure. We also use a coverage metric based on the mean dispersion of failure trajectories in disturbance space~\citep{esposito2004adaptive}. The mean dispersion  $C_{disp} \in [0,1]$ is computed over a grid in trajectory space as
\begin{equation*}
    C_{disp} = 1 - \frac{1}{n} \sum_{j=1}^{n} \frac{\min (d_j, g)}{g}
\end{equation*}
where $n$ is the number of grid points, $g$ is the grid spacing, and $d_j$ is the minimum distance from the grid point to a point in the sampled disturbance trajectories. Larger values of $C_{disp}$ indicate that a sampler has discovered failures in more regions of disturbance trajectory-space.

We compare the proposed gradient-based approach to direct Monte Carlo (MC) sampling and the black-box MCMC Particle Gibbs (PG)~\citep{andrieu2010particle} algorithm with \num{1000} particles.


\subsection{Inverted Pendulum}

The first system we validate is an inverted pendulum with a neural network controller. While this is not a safety-critical system, it demonstrates the ability of the proposed approach to validate policies with machine learning components. In the inverted pendulum, a policy exerts a control torque to keep the pendulum balanced upright. The state of the system is $s=(\theta, \dot{\theta})$ where $\theta$ is the angle of the pendulum from vertical and $\dot{\theta}$ is the angular velocity. The control policy takes continuous-valued actions $a \in [-T_{\max}, T_{\max}]$, where $T_{\max}$ is the maximum control effort of the actuator. The pendulum is underactuated, meaning that the available torque is too weak to result in arbitrary state trajectories. If the pendulum reaches beyond about $\SI{0.5}{\radian}$ from vertical, the available torque will not be able to overcome the torque due to the pendulum's mass, and the pendulum will fall. The pendulum and failure condition is illustrated in \cref{fig:systems-pendulum}.

We optimize a neural network policy using the Proximal Policy Optimization~\citep{schulman2017ppo} reinforcement learning algorithm, which has been shown to perform well on the inverted pendulum. The policy is represented by a two-layer neural network with \num{32} neurons per layer and tanh activations between hidden layers. The policy is trained for \SI{e6} time steps. 

We sample failures of control policy by considering external torque disturbances from the environment. Adversarial torque disturbances are assumed to be distributed according to a zero-mean Gaussian with variance $\sigma_n$. This experiment uses $\sigma_n = \SI{0.1}{\newton \meter}$ and $T_{\max}=\SI{2}{\newton \meter}$ so that disturbances are likely to be much smaller than the maximum control torque. A failure occurs when the torque due to gravity overcomes the maximum control torque, which occurs when $| \theta | \geq \SI{0.5}{\radian}$. The distance to failure is the minimum distance from the failure threshold. Simulations span a $\SI{5}{\second}$ horizon with a $\SI{0.1}{\second}$ time step. We sample \num{10000} samples for each method, with \num{10} MCMC chains and \num{1000} samples per chain.


\subsection{Autonomous Vehicle Scenario}
\begin{figure}[t!]
    \label{fig:systems}
    \subfigure[\small Inverted Pendulum]{\label{fig:systems-pendulum} \includegraphics[width=0.25\linewidth]{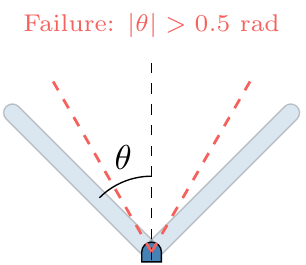}}
    \hfill
    \subfigure[\small AV Pedestrian Scenario]{\label{fig:systems-jaywalk} \raisebox{0.25\height} {\begin{tikzpicture}[x=0.04cm,y=-0.04cm, dot/.style = {circle, fill, minimum size=#1,
              inner sep=0pt, outer sep=0pt},
dot/.default = 1.2pt  
                    ]

  \begin{scope}

    \clip [rounded corners=3] (0, -5) rectangle (60, 29);

    \fill [gray!50] (0,-5) rectangle (60,29);
    \fill [green!40!black!40] (0, -5) rectangle (60, 0); 
    \fill [green!40!black!40] (0, 16) rectangle (60, 29);  

    \draw [white, dashed, thick] (0,8) -- (60,8);  

    


    \node[sedan top, body color=steelblue, window color=black!80,minimum width=0.5cm,](sedan) at (8, 12){};
    \node[circle, fill=pastelred, inner sep=0pt, minimum size=3pt](ped) at (40, 22){};
    \node[](sedanarrow) at (25, 12){};
    \node[](sedanstart) at (11, 12){};
    \draw [->, >=stealth] (sedanstart) -- (sedanarrow);

    \node[](pedarrow) at (40, 13){};
    \draw [->, >=stealth] (ped) edge (pedarrow);

  \end{scope}

\end{tikzpicture}}}
    \hfill
    \subfigure[\small Lunar Lander]{\label{fig:systems-lander}  
\begin{tikzpicture}[ one end extended/.style={shorten >=-#1},
 one end extended/.default=5cm,]

    \node[rectangle, fill=black!20, minimum width=10cm, minimum height=1cm] at (0,-5.5){};

    \node[trapezium,draw, rotate=-20, minimum height=1.2cm, text width=0.7, trapezium angle=75, fill=cyan!40!black] (t) at (0,0) {};
    
    \draw [add=0 and 3.9, dashed, color=black!] (t.north) to node[left] {\LARGE $AGL$} (t.south);
    
    \node[trapezium,draw, rotate=-20, minimum height=1.2cm, text width=0.7, trapezium angle=75, fill=cyan!50!black!60] (t) at (0,0) {};
    
    \draw [solid, color=black!, thick] (-5, -5) to (5, -5);
    \draw [solid, color=pastelred, thick, line width=1mm] (-3, -5) -- node[below] {\LARGE Landing Goal} (3, -5);
    \node[](ttop) at (t.north west){};
    \node [position=75:{15mm} from t] (tpoint) {};
    
    \draw [arrow] (t) edge  node[right] {\LARGE $T$} (tpoint) {};
    
    \node [position=160:{15mm} from ttop] (fpoint) {};
    
    \draw [arrow] (t.north west) edge  node[below] {\LARGE $F_x$} (fpoint) {};
    
    \node[] (tsw) at (t.bottom left corner) {};
    \node [position=235:{5mm} from tsw] (leg1) {};
    \draw [solid, color=black!80, thick, line width = 1mm] (t.bottom left corner)++(0.0mm,0.0mm) to (leg1);
    
    \node[] (tse) at (t.bottom right corner) {};
    \node [position=270:{5mm} from tse] (leg2) {};
    \draw [solid, color=black!80, thick, line width = 1mm] (t.bottom right corner)++(0.0mm, 0.0mm) to (leg2);

    \node[trapezium, draw=black!80, rotate=-20, minimum height=1.2cm, text width=0.7, trapezium angle=75, fill=steelblue] (t) at (0,0) {};

    \node[circle, fill=black!80, inner sep=0pt, minimum size=3pt] at (t.center){};
    \draw[ densely dotted] (t.center) edge node[left] {\LARGE $d$} (t.north){};

\end{tikzpicture}
 }
    \caption{The three environments used in validation experiments.}
\end{figure}
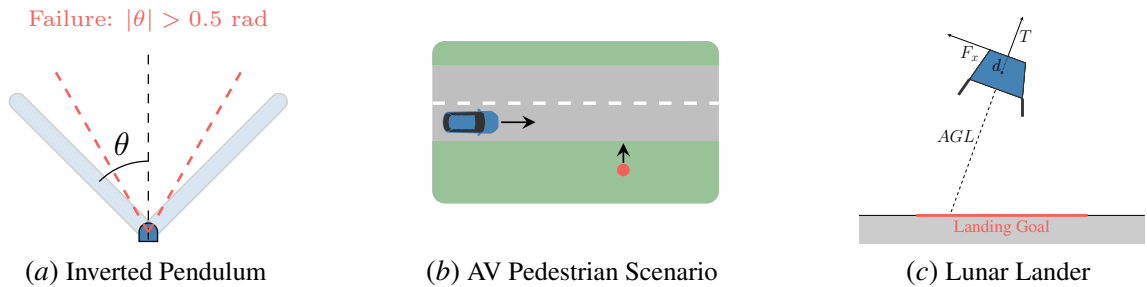

The second experiment involves an autonomous vehicle (AV) approaching a single pedestrian at a crosswalk and is based on the work of \citet{koren2018adaptive}. This scenario is illustrated in \cref{fig:systems-jaywalk}. The system under test is the autonomous vehicle's control policy, which is defined by the Intelligent Driver Model (IDM) \citep{treiber2000IDM}. The IDM is a model-based controller that keeps the vehicle in its lane, following the traffic or other obstacles ahead while maintaining a safe distance.  The goal of the AV is to identify the
pedestrian crossing the road and apply adequate braking to avoid coming too close. Each agent's state is represented by a 4-tuple $s = (r_x, r_y, v_x, v_y)$ where $(r_x, r_y)$ is the agent's position and $(v_x, v_y)$ is the agent's velocity. In the noise-free case, the AV comes to a stop and the pedestrian safely crosses.

The adversary has control over pedestrian acceleration and noise in the AV's sensors. The disturbances to the system are represented by a 6-tuple $x = (\delta r_x, \delta r_y, \delta v_x, \delta v_y, a_x, a_y)$ where $(\delta r_x, \delta r_y)$ is the noise in the pedestrian's observed position, $(\delta v_x, \delta v_y)$ is the noise in the pedestrian's observed velocity, and $(a_x, a_y)$ is the pedestrian's acceleration. The noise components are processed at each time step and added to the vehicle’s observation of each pedestrian’s position and velocity.
The disturbances are assumed to be distributed according to a multivariate Gaussian distribution with zero mean and diagonal covariance matrix. The diagonal components of covariance are $\sigma^2_{r}=0.1$ for position measurement noise, $\sigma^2_{v}=0.1$ for velocity measurement noise, $\sigma^2_{lat}=0.01$ over the pedestrian's lateral acceleration, and $\sigma^2_{lon}=0.1$ over the pedestrian's longitudinal acceleration. These values are chosen to reflect relatively small sensor noise, and that the pedestrian is less likely to accelerate longitudinally while crossing the street.

Failures of this scenario are defined by any collision between the pedestrian and vehicle. The distance to failure is the minimum Euclidean distance between the vehicle and the pedestrian. We sample \num{20000} samples for each method, with \num{20} MCMC chains and \num{1000} samples per chain.

\subsection{Partially Observable Lunar lander}
The final validation problem is a simulation of a lunar lander with partially observed state information. The objective of the lunar lander's policy is to guide the vehicle in a target area with a soft landing. The vehicle state is represented by the 6-tuple $(x, y, \theta, \dot{x}, \dot{y}, \dot{\theta})$, where $x$ is the horizontal position, $y$ is the vertical position, and $\theta$ is the vehicle's orientation. The vehicle's dynamics are deterministic. The vehicle makes noisy observations of its angular rate $\omega$, horizontal speed $v$, and above ground level altitude ($AGL$)\@. $AGL$ is a noisy measurement of the distance from the vehicle's center of mass to the ground along the vehicle's longitudinal axis. The continuous actions are defined by $(T, F_x, d)$ where $T$ is the main thrust acting along the longitudinal axis, $F_x$ is lateral corrective thrust acting at an offset of $d$ from the center of mass. The lander is illustrated in \cref{fig:systems-lander}.

We optimize a control policy by considering a discrete, fully observable version of the continuous, partially observable problem. We use approximate value iteration to solve for a table-based policy in the discrete problem. Next, we create a differentiable representation of the table-based policy by performing behavior cloning with a neural network approximator. The network is trained to minimize the mean squared error between the continuous output and the table-based action. 

We assume that observation noise is identically and independently distributed at each time step according to a zero-mean Gaussian with a diagonal covariance matrix. We use $\sigma^2_{\omega}=0.02$,   $\sigma^2_{v}=0.1$, and $\sigma^2_{AGL}=1.0$.  An extended Kalman filter maintains a multivariate Gaussian belief over the true state in the partially observable problem. The policy uses the mean to calculate an action.

Our approach samples sequences of observation noise that lead to a hard landing, defined as the landing velocity being greater than \SI{6}{\meter\per\second}. Empirically, we discovered the lunar lander contained multiple isolated failure modes. We select initial conditions for the MCMC samplers by performing black-box optimization of the probabilistic model's posterior density. Specifically, we use particle swarm optimization with five particles and \num{100} iterations to select initial conditions for each MCMC chain. We sample \num{30} MCMC chains with \num{1000} samples each to explore the different failure modes.

\section{Results}\label{sec:results}
This section presents experimental results for each validation problem. Probabilistic models of each system were written using Turing and sampled using the HMC-NUTS implementation provided by \citet{ge2018turing}. Gradients were computed using the Zygote \citep{innes2019zygote} AD framework. Experiments were ran in serial on a desktop with 32GB of RAM and an Intel i7-7700K CPU\@.

\ph{Overview}
\cref{4} shows the failure rate, failure trajectory log-likelihood, sampling time per failure, and mean dispersion for each sampling method and validation setting. The proposed approach consistently outperforms baselines in terms of failure rate. The gradient-based HMC is able to consistently sample from the high-dimensional failure region, while PG and MC samplers struggle. Additionally, the results suggest that HMC consistently samples from the high-likelihood failure region. The computational cost of computing gradients through the state trajectories is outweighed by the increase in sampling performance. HMC samples failures \num{5}--\num{10} times faster in wall clock time compared to baselines. Failures sampled using HMC also cover a significantly greater portion of the disturbance trajectory space across all experiments. Greater disturbance space coverage generally corresponds to a more diverse set of failures being sampled.

\begin{table*}[t!]\label{table:all-results}
\sisetup{
         table-text-alignment=right,
        }
    \begin{center}
    \caption{Results for the inverted pendulum (IP), autonomous vehicle scenario (AV), and partially observable lunar lander (POL). Failure rate,  trajectory mean and maximum log-likelihood (LL), sampling time per failure, mean dispersion $C_{disp}$, and total sampling time are reported for each algorithm. HMC outperforms black-box baselines in terms of failure rate, sampling time per failure, and disturbance-space coverage.}
        \resizebox{\textwidth}{!}{
        \begin{tabular}{@{}l l
        S[table-format=1.2e-1, scientific-notation = true]
        S[table-format=3.2(1), round-precision=4, separate-uncertainty=true, table-align-uncertainty]
        S[table-format=3.1, round-precision=1]
        S[ table-text-alignment=left, table-format=1e1, round-mode=figures, round-precision=2, scientific-notation = true]
        S[ table-text-alignment=left, table-format=1e1, round-mode=figures, round-precision=2, scientific-notation = true]
        S[round-mode=places, round-precision=0]
        @{}}
        \toprule
         & Method & $\text{Failure Rate}$ & $\text{Mean LL}$ & $\text{Max. LL}$ & $\text{Failures} / \si{\s}$ & $C_{disp}$ & $\text{Time (\si{\min})}$\\ 
        \midrule
        
        \multirow{3}{*}{\rotatebox[origin=c]{90}{IP}} & \textbf{HMC} & \num{9.8e-1} & \num{-24.66(959)} & \num{-8.18} & \num{8.0} & \num{0.75} & \num{22}\\ 
        & PG & \num{2.0e-4} & \num{-21.42(15)} & \num{-19.89} & \num{1.05e-3} & \num{0.00021} & \num{27}\\
        & MC & \num{3.0e-3} & \num{-18.44(297)} & \num{-13.81} & \num{1.22} & \num{0.024} & \num{2}\\ 
        
        \midrule
        
        \multirow{3}{*}{\rotatebox[origin=c]{90}{AV}} & \textbf{HMC} & \num{7.8e-1}  & \num{-2.01(2000)} & \num{0.53} & \num{7.81} & \num{0.020} & \num{40}\\ 
        & PG & \num{1.7e-1} & \num{-0.71(48)} & \num{0.11} & \num{6.5e-3} & \num{1.0e-5} & \num{40}\\
        & MC & \num{5.0e-4} & \num{-12.05(6)}  & \num{-12.35} & \num{0.85} & \num{1.0e-9} & \num{3}\\ 
        
        \midrule
        
        \multirow{3}{*}{\rotatebox[origin=c]{90}{POL}} & \textbf{HMC} & \num{1.6e-1}  & \num{1.68(169)} & \num{3.19} & \num{0.41} & \num{0.21} & \num{375}\\ 
        & PG & \num{2.4e-4} & \num{1.80(021)} & \num{2.12} & \num{3.3e-5} & \num{3.2e-3} & \num{810}\\
        & MC & \num{3.0e-5} & \num{2.34(109)} & \num{2.61} & \num{0.17} & \num{1.5e-5} & \num{13}\\ 
        
        \bottomrule
        \end{tabular}}
        \label{4}
    \end{center}
\end{table*}

\begin{figure}[t!]
    \subfigure[\small HMC]{\includestandalone[width=0.32\linewidth]{figures/pendulum_failures_hmc}}
    \hfill
    \subfigure[\small PG]{
\usetikzlibrary{arrows.meta}
\usetikzlibrary{backgrounds}
\usepgfplotslibrary{patchplots}
\usepgfplotslibrary{fillbetween}
\pgfplotsset{%
    layers/standard/.define layer set={%
        background,axis background,axis grid,axis ticks,axis lines,axis tick labels,pre main,main,axis descriptions,axis foreground%
    }{
        grid style={/pgfplots/on layer=axis grid},%
        tick style={/pgfplots/on layer=axis ticks},%
        axis line style={/pgfplots/on layer=axis lines},%
        label style={/pgfplots/on layer=axis descriptions},%
        legend style={/pgfplots/on layer=axis descriptions},%
        title style={/pgfplots/on layer=axis descriptions},%
        colorbar style={/pgfplots/on layer=axis descriptions},%
        ticklabel style={/pgfplots/on layer=axis tick labels},%
        axis background@ style={/pgfplots/on layer=axis background},%
        3d box foreground style={/pgfplots/on layer=axis foreground},%
    },
    compat=1.16
}

\begin{tikzpicture}[/tikz/background rectangle/.style={fill={rgb,1:red,1.0;green,1.0;blue,1.0}, draw opacity={1.0}}, show background rectangle]
    \begin{axis}[point meta max={nan}, point meta min={nan}, legend cell align={left}, legend columns={1}, title={}, title style={at={{(0.5,1)}}, anchor={south}, font={{\fontsize{14 pt}{18.2 pt}\selectfont}}, color={rgb,1:red,0.0;green,0.0;blue,0.0}, draw opacity={1.0}, rotate={0.0}}, legend style={color={rgb,1:red,0.0;green,0.0;blue,0.0}, draw opacity={1.0}, line width={1}, solid, fill={rgb,1:red,1.0;green,1.0;blue,1.0}, fill opacity={1.0}, text opacity={1.0}, font={{\fontsize{36 pt}{46.800000000000004 pt}\selectfont}}, text={rgb,1:red,0.0;green,0.0;blue,0.0}, cells={anchor={center}}, at={(0.02, 0.98)}, anchor={north west}}, axis background/.style={fill={rgb,1:red,1.0;green,1.0;blue,1.0}, opacity={1.0}}, anchor={north west}, xshift={1.0mm}, yshift={-1.0mm}, width={150.4mm}, height={99.6mm}, scaled x ticks={false}, xlabel={time step}, x tick style={color={rgb,1:red,0.0;green,0.0;blue,0.0}, opacity={1.0}}, x tick label style={color={rgb,1:red,0.0;green,0.0;blue,0.0}, opacity={1.0}, rotate={0}}, xlabel style={at={(ticklabel cs:0.5)}, anchor=near ticklabel, at={{(ticklabel cs:0.5)}}, anchor={near ticklabel}, font={{\fontsize{36 pt}{46.800000000000004 pt}\selectfont}}, color={rgb,1:red,0.0;green,0.0;blue,0.0}, draw opacity={1.0}, rotate={0.0}}, xmajorgrids={true}, xmin={0}, xmax={50}, xticklabels={{$0$,$10$,$20$,$30$,$40$,$50$}}, xtick={{0.0,10.0,20.0,30.0,40.0,50.0}}, xtick align={inside}, xticklabel style={font={{\fontsize{36 pt}{46.800000000000004 pt}\selectfont}}, color={rgb,1:red,0.0;green,0.0;blue,0.0}, draw opacity={1.0}, rotate={0.0}}, x grid style={color={rgb,1:red,0.0;green,0.0;blue,0.0}, draw opacity={0.1}, line width={0.5}, solid}, axis x line*={left}, x axis line style={color={rgb,1:red,0.0;green,0.0;blue,0.0}, draw opacity={1.0}, line width={1}, solid}, scaled y ticks={false}, ylabel={$\theta$ (rad)}, y tick style={color={rgb,1:red,0.0;green,0.0;blue,0.0}, opacity={1.0}}, y tick label style={color={rgb,1:red,0.0;green,0.0;blue,0.0}, opacity={1.0}, rotate={0}}, ylabel style={at={(ticklabel cs:0.5)}, anchor=near ticklabel, at={{(ticklabel cs:0.5)}}, anchor={near ticklabel}, font={{\fontsize{36 pt}{46.800000000000004 pt}\selectfont}}, color={rgb,1:red,0.0;green,0.0;blue,0.0}, draw opacity={1.0}, rotate={0.0}}, ymajorgrids={true}, ymin={-0.75}, ymax={0.75}, yticklabels={{$-0.75$,$-0.50$,$-0.25$,$0.00$,$0.25$,$0.50$,$0.75$}}, ytick={{-0.75,-0.5,-0.25,0.0,0.25,0.5,0.75}}, ytick align={inside}, yticklabel style={font={{\fontsize{36 pt}{46.800000000000004 pt}\selectfont}}, color={rgb,1:red,0.0;green,0.0;blue,0.0}, draw opacity={1.0}, rotate={0.0}}, y grid style={color={rgb,1:red,0.0;green,0.0;blue,0.0}, draw opacity={0.1}, line width={0.5}, solid}, axis y line*={left}, y axis line style={color={rgb,1:red,0.0;green,0.0;blue,0.0}, draw opacity={1.0}, line width={1}, solid}, colorbar={false}]
        \addplot[color={rgb,1:red,1.0;green,0.0;blue,0.0}, name path={92058058-91eb-4cfb-b4a8-f2bef7e27e65}, area legend, fill={rgb,1:red,1.0;green,0.0;blue,0.0}, fill opacity={0.5}, draw opacity={0.5}, line width={0}, solid]
            table[row sep={\\}]
            {
                \\
                0.0  0.5  \\
                50.0  0.5  \\
                50.0  1.0  \\
                0.0  1.0  \\
            }
            ;
        \addlegendentry {$s \notin \psi$}
        \addplot[color={rgb,1:red,1.0;green,0.0;blue,0.0}, name path={7edfad6b-a2a1-4d1f-99d7-a66775b993c4}, area legend, fill={rgb,1:red,1.0;green,0.0;blue,0.0}, fill opacity={0.5}, draw opacity={0.5}, line width={0}, solid, forget plot]
            table[row sep={\\}]
            {
                \\
                0.0  -1.0  \\
                50.0  -1.0  \\
                50.0  -0.5  \\
                0.0  -0.5  \\
            }
            ;
        \addplot[color={rgb,1:red,0.2745;green,0.5098;blue,0.7059}, name path={fdeecc74-76b3-4ad9-bc05-3e00442f5566}, draw opacity={0.8}, line width={1}, solid, forget plot]
            table[row sep={\\}]
            {
                \\
                1.0  0.0  \\
                2.0  0.006565826926237105  \\
                3.0  -0.0013026911944186148  \\
                4.0  -0.018006012181354784  \\
                5.0  -0.024859006884800236  \\
                6.0  -0.03396935581874932  \\
                7.0  -0.02967839890493447  \\
                8.0  -0.014609360178446384  \\
                9.0  -0.010622369718646878  \\
                10.0  0.0025325028879148413  \\
                11.0  0.02435961550860845  \\
                12.0  0.03756214131486883  \\
                13.0  0.039750043806911384  \\
                14.0  0.036882833996017494  \\
                15.0  0.019304019886378764  \\
                16.0  0.014891701045108844  \\
                17.0  0.007802296565013323  \\
                18.0  0.004177604969655692  \\
                19.0  0.007798793375474603  \\
                20.0  0.024489389776760455  \\
                21.0  0.0383457835249281  \\
                22.0  0.05412531158691562  \\
                23.0  0.057996149632532294  \\
                24.0  0.038837709041522346  \\
                25.0  0.04439508837408329  \\
                26.0  0.03583076804092342  \\
                27.0  0.03766121017172606  \\
                28.0  0.029595847680442308  \\
                29.0  0.01698878442036604  \\
                30.0  0.011169641537609888  \\
                31.0  0.004262737781488196  \\
                32.0  -0.006417602117985994  \\
                33.0  -0.036281002080235145  \\
                34.0  -0.06213983906826719  \\
                35.0  -0.10289288701877583  \\
                36.0  -0.14761259563009865  \\
                37.0  -0.18873584788405656  \\
                38.0  -0.2458384750157256  \\
                39.0  -0.2970129093138565  \\
                40.0  -0.36651295535151374  \\
                41.0  -0.44630770797860575  \\
                42.0  -0.5591809970770796  \\
            }
            ;
    \end{axis}
    \end{tikzpicture}}
    \hfill
    \subfigure[\small MC]{\includestandalone[width=0.32\linewidth]{figures/pendulum_failures_mc}}
    \caption{Failure trajectories for the inverted pendulum with HMC and baseline methods. HMC discovers many likely failures in the positive and negative directions. Greater opacity reflects a higher likelihood failure.}
    \label{fig:pendulum-failures}
\end{figure}
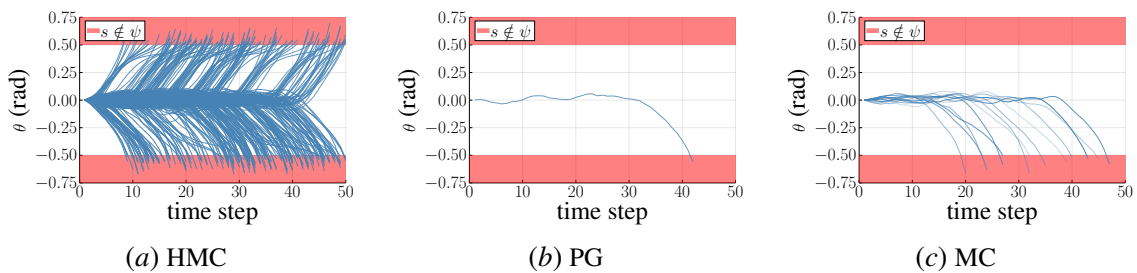

\ph{Inverted Pendulum}
\cref{fig:pendulum-failures} shows failures sampled using the HMC and baseline methods. The number of failure trajectories for HMC is downsampled for clarity. HMC finds a more diverse set of failure modes compared to PG and MC, with many failures in the positive- and negative- $\theta$ direction. The failures found by HMC span a wide range of trajectory lengths and failure modes. HMC uses the gradient of the trajectory with respect to both the pendulum dynamics and the neural network controller, enabling much broader coverage of the failure distribution than black-box baselines.

HMC samples cover the disturbance space well and uncover asymmetric failure modes. The most likely failure trajectories found with each method lead to the pendulum falling in the negative-$\theta$ direction. HMC failure samples indicate that the neural network controller exhibits different failure modes in the positive- and negative-$\theta$ directions. Positive-$\theta$ failures have a lower mean log-likelihood of \num{-28.1}, while negative-$\theta$ failures have a mean of \num{-17.4}. The pendulum tends to fall more directly in the negative direction, while some failures in the positive direction hesitate before falling. This may indicate that this particular controller learned by PPO is more robust to disturbances in the positive-$\theta$ direction than in the negative-$\theta$ direction.

\begin{figure}[t!]
    \subfigure[\small HMC]{\includestandalone[width=0.31\linewidth]{crosswalk2_failures_hmc}}
    \hfill
    \subfigure[\small PG]{\includestandalone[width=0.31\linewidth]{crosswalk2_failures_pg}}
    \hfill
    \subfigure[\small MC]{
\usetikzlibrary{arrows.meta}
\usetikzlibrary{backgrounds}
\usepgfplotslibrary{patchplots}
\usepgfplotslibrary{fillbetween}
\pgfplotsset{%
    layers/standard/.define layer set={%
        background,axis background,axis grid,axis ticks,axis lines,axis tick labels,pre main,main,axis descriptions,axis foreground%
    }{
        grid style={/pgfplots/on layer=axis grid},%
        tick style={/pgfplots/on layer=axis ticks},%
        axis line style={/pgfplots/on layer=axis lines},%
        label style={/pgfplots/on layer=axis descriptions},%
        legend style={/pgfplots/on layer=axis descriptions},%
        title style={/pgfplots/on layer=axis descriptions},%
        colorbar style={/pgfplots/on layer=axis descriptions},%
        ticklabel style={/pgfplots/on layer=axis tick labels},%
        axis background@ style={/pgfplots/on layer=axis background},%
        3d box foreground style={/pgfplots/on layer=axis foreground},%
    },
    compat=1.16,
}

\begin{tikzpicture}[/tikz/background rectangle/.style={fill={rgb,1:red,1.0;green,1.0;blue,1.0}, draw opacity={1.0}}, show background rectangle]
\begin{axis}[point meta max={nan}, point meta min={nan}, legend cell align={left}, legend columns={1}, title={}, title style={at={{(0.5,1)}}, anchor={south}, font={{\fontsize{14 pt}{18.2 pt}\selectfont}}, color={rgb,1:red,0.0;green,0.0;blue,0.0}, draw opacity={1.0}, rotate={0.0}}, legend style={color={rgb,1:red,0.0;green,0.0;blue,0.0}, draw opacity={1.0}, line width={1}, solid, fill={rgb,1:red,1.0;green,1.0;blue,1.0}, fill opacity={1.0}, text opacity={1.0}, font={{\fontsize{36 pt}{46.800000000000004 pt}\selectfont}}, text={rgb,1:red,0.0;green,0.0;blue,0.0}, cells={anchor={center}}, at={(1.02, 1)}, anchor={north west}}, axis background/.style={fill={rgb,1:red,1.0;green,1.0;blue,1.0}, opacity={1.0}}, anchor={north west}, xshift={1.0mm}, yshift={-1.0mm}, width={145.4mm}, height={99.6mm}, scaled x ticks={false}, xlabel={$r_x$ (m)}, x tick style={color={rgb,1:red,0.0;green,0.0;blue,0.0}, opacity={1.0}}, x tick label style={color={rgb,1:red,0.0;green,0.0;blue,0.0}, opacity={1.0}, rotate={0}}, xlabel style={at={(ticklabel cs:0.5)}, anchor=near ticklabel, at={{(ticklabel cs:0.5)}}, anchor={near ticklabel}, font={{\fontsize{36 pt}{46.800000000000004 pt}\selectfont}}, color={rgb,1:red,0.0;green,0.0;blue,0.0}, draw opacity={1.0}, rotate={0.0}}, xmajorgrids={true}, xmin={-4.0}, xmax={0.25}, xticklabels={{$-4$,$-3$,$-2$,$-1$,$0$}}, xtick={{-4.0,-3.0,-2.0,-1.0,0.0}}, xtick align={inside}, xticklabel style={font={{\fontsize{36 pt}{46.800000000000004 pt}\selectfont}}, color={rgb,1:red,0.0;green,0.0;blue,0.0}, draw opacity={1.0}, rotate={0.0}}, x grid style={color={rgb,1:red,0.0;green,0.0;blue,0.0}, draw opacity={0.1}, line width={0.5}, solid}, axis x line*={left}, x axis line style={color={rgb,1:red,0.0;green,0.0;blue,0.0}, draw opacity={1.0}, line width={1}, solid}, scaled y ticks={false}, ylabel={$r_y$ (m)}, y tick style={color={rgb,1:red,0.0;green,0.0;blue,0.0}, opacity={1.0}}, y tick label style={color={rgb,1:red,0.0;green,0.0;blue,0.0}, opacity={1.0}, rotate={0}}, ylabel style={at={(ticklabel cs:0.5)}, anchor=near ticklabel, at={{(ticklabel cs:0.5)}}, anchor={near ticklabel}, font={{\fontsize{36 pt}{46.800000000000004 pt}\selectfont}}, color={rgb,1:red,0.0;green,0.0;blue,0.0}, draw opacity={1.0}, rotate={0.0}}, ymajorgrids={true}, ymin={-2.0684064935887516}, ymax={0.34862294654714504}, yticklabels={{$-2.0$,$-1.5$,$-1.0$,$-0.5$,$0.0$}}, ytick={{-2.0,-1.5,-1.0,-0.5,0.0}}, ytick align={inside}, yticklabel style={font={{\fontsize{36 pt}{46.800000000000004 pt}\selectfont}}, color={rgb,1:red,0.0;green,0.0;blue,0.0}, draw opacity={1.0}, rotate={0.0}}, y grid style={color={rgb,1:red,0.0;green,0.0;blue,0.0}, draw opacity={0.1}, line width={0.5}, solid}, axis y line*={left}, y axis line style={color={rgb,1:red,0.0;green,0.0;blue,0.0}, draw opacity={1.0}, line width={1}, solid}, colorbar={false}]
    \addplot[color={rgb,1:red,0.2745;green,0.5098;blue,0.7059}, name path={aa561241-359b-4918-b6d8-3a222a72ad4e}, draw opacity={0.9}, line width={1}, solid, forget plot]
        table[row sep={\\}]
        {
            \\
            0.0  -2.0  \\
            0.0  -1.8399999999999999  \\
            -0.0002917533878661254  -1.6780970195508809  \\
            0.0006616020528376042  -1.517824494543224  \\
            0.0020809297451362552  -1.3635991317257778  \\
            0.0033375342494528257  -1.2078599134724368  \\
            0.004012040386997051  -1.0545813650068987  \\
            0.004496207390519823  -0.9054429722156003  \\
            0.004548826373025369  -0.7637214712756184  \\
            0.003988249335032465  -0.6278550083340224  \\
            0.0037486535640294384  -0.49437714247159437  \\
            0.0024680380501313093  -0.36467035589957836  \\
            0.00217524096366306  -0.2330906377822522  \\
            0.0013572731515419047  -0.10454377247896829  \\
            0.00028062731471987373  0.022919669198113418  \\
            0.0013659505542897843  0.1520266399626989  \\
            0.003349066665991539  0.28021645295839326  \\
        }
        ;
\end{axis}
\end{tikzpicture}
    }
    \caption{Pedestrian trajectories resulting in collision for the autonomous vehicle scenario with HMC and baseline methods. HMC samples failures with a mixture of high sensor noise and lateral motion. Baseline methods only sample from the sensor noise failure mode. Greater opacity reflects a higher likelihood failure.}
    \label{fig:avcollisions}
\end{figure}
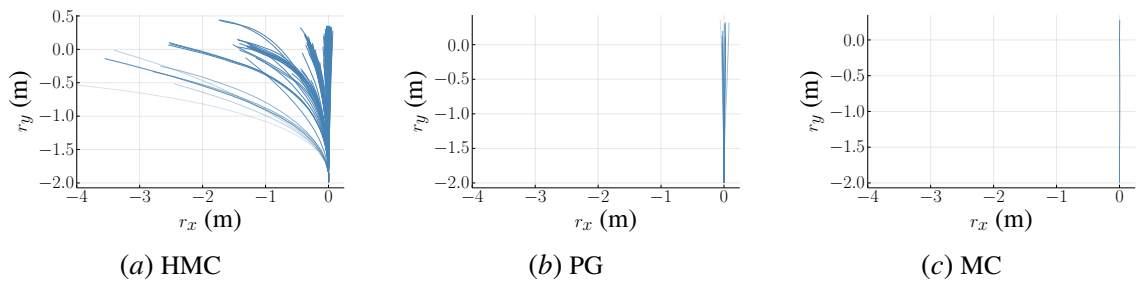

\ph{Autonomous Vehicle Scenario}
Sampled pedestrian trajectories that result in a collision are shown in \cref{fig:avcollisions}. The HMC samples are subsampled to $500$ trajectories for clarity. All trajectories end at the time of collision. HMC finds a wide variety of failure modes in terms of the trajectory of the pedestrian while the baselines focus on a single mode.

The more likely failure modes involve little change in the acceleration of the pedestrian. These failure trajectories are concentrated towards the $y$-axis, and introduce sensor noise to cause the AV to incorrectly estimate the relative distance or velocity to the pedestrian. Other failure modes involve accelerations of the pedestrian in the negative-$x$ direction. Baseline methods were unable to sample from these modes. In extreme cases, the AV comes to a complete stop in front of the crosswalk and the pedestrian walks into the vehicle. This failure mode has been observed in previous work~\citep{koren2018adaptive, corso2019adaptive}. While this pedestrian-induced failure might not be crucial to the safety of the AV policy, it highlights important questions for system designers such as accident blame that are not apparent from the baseline results.

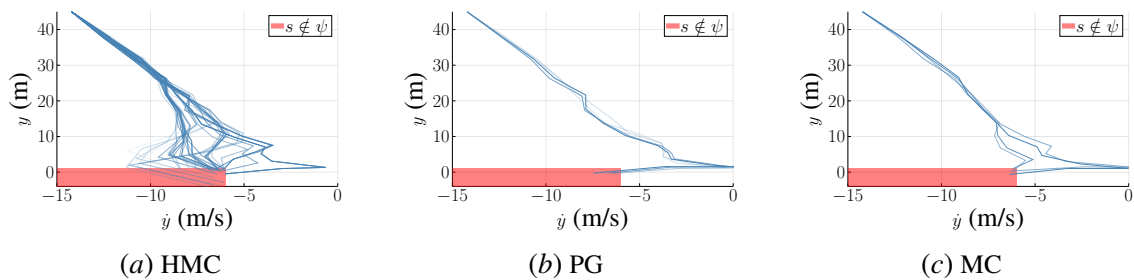
\begin{figure}[t!]
    \subfigure[\small HMC]{\includestandalone[width=0.31\linewidth]{lander_failures_hmc}}
    \hfill
    \subfigure[\small PG]{\includestandalone[width=0.31\linewidth]{lander_failures_pg}}
    \hfill
    \subfigure[\small MC]{    
\usetikzlibrary{arrows.meta}
\usetikzlibrary{backgrounds}
\usepgfplotslibrary{patchplots}
\usepgfplotslibrary{fillbetween}
\pgfplotsset{%
    layers/standard/.define layer set={%
        background,axis background,axis grid,axis ticks,axis lines,axis tick labels,pre main,main,axis descriptions,axis foreground%
    }{
        grid style={/pgfplots/on layer=axis grid},%
        tick style={/pgfplots/on layer=axis ticks},%
        axis line style={/pgfplots/on layer=axis lines},%
        label style={/pgfplots/on layer=axis descriptions},%
        legend style={/pgfplots/on layer=axis descriptions},%
        title style={/pgfplots/on layer=axis descriptions},%
        colorbar style={/pgfplots/on layer=axis descriptions},%
        ticklabel style={/pgfplots/on layer=axis tick labels},%
        axis background@ style={/pgfplots/on layer=axis background},%
        3d box foreground style={/pgfplots/on layer=axis foreground},%
    },
    compat=1.16,
}

\begin{tikzpicture}[/tikz/background rectangle/.style={fill={rgb,1:red,1.0;green,1.0;blue,1.0}, draw opacity={1.0}}, show background rectangle]
    \begin{axis}[point meta max={nan}, point meta min={nan}, legend cell align={left}, legend columns={1}, title={}, title style={at={{(0.5,1)}}, anchor={south}, font={{\fontsize{14 pt}{18.2 pt}\selectfont}}, color={rgb,1:red,0.0;green,0.0;blue,0.0}, draw opacity={1.0}, rotate={0.0}}, legend style={color={rgb,1:red,0.0;green,0.0;blue,0.0}, draw opacity={1.0}, line width={1}, solid, fill={rgb,1:red,1.0;green,1.0;blue,1.0}, fill opacity={1.0}, text opacity={1.0}, font={{\fontsize{36 pt}{46.800000000000004 pt}\selectfont}}, text={rgb,1:red,0.0;green,0.0;blue,0.0}, cells={anchor={center}}, at={(0.98, 0.98)}, anchor={north east}}, axis background/.style={fill={rgb,1:red,1.0;green,1.0;blue,1.0}, opacity={1.0}}, anchor={north west}, xshift={1.0mm}, yshift={-1.0mm}, width={150.4mm}, height={99.6mm}, scaled x ticks={false}, xlabel={$\dot{y}$ (m/s)}, x tick style={color={rgb,1:red,0.0;green,0.0;blue,0.0}, opacity={1.0}}, x tick label style={color={rgb,1:red,0.0;green,0.0;blue,0.0}, opacity={1.0}, rotate={0}}, xlabel style={at={(ticklabel cs:0.5)}, anchor=near ticklabel, at={{(ticklabel cs:0.5)}}, anchor={near ticklabel}, font={{\fontsize{36 pt}{46.800000000000004 pt}\selectfont}}, color={rgb,1:red,0.0;green,0.0;blue,0.0}, draw opacity={1.0}, rotate={0.0}}, xmajorgrids={true}, xmin={-15}, xmax={0}, xticklabels={{$-15$,$-10$,$-5$,$0$}}, xtick={{-15.0,-10.0,-5.0,0.0}}, xtick align={inside}, xticklabel style={font={{\fontsize{36 pt}{46.800000000000004 pt}\selectfont}}, color={rgb,1:red,0.0;green,0.0;blue,0.0}, draw opacity={1.0}, rotate={0.0}}, x grid style={color={rgb,1:red,0.0;green,0.0;blue,0.0}, draw opacity={0.1}, line width={0.5}, solid}, axis x line*={left}, x axis line style={color={rgb,1:red,0.0;green,0.0;blue,0.0}, draw opacity={1.0}, line width={1}, solid}, scaled y ticks={false}, ylabel={$y$ (m)}, y tick style={color={rgb,1:red,0.0;green,0.0;blue,0.0}, opacity={1.0}}, y tick label style={color={rgb,1:red,0.0;green,0.0;blue,0.0}, opacity={1.0}, rotate={0}}, ylabel style={at={(ticklabel cs:0.5)}, anchor=near ticklabel, at={{(ticklabel cs:0.5)}}, anchor={near ticklabel}, font={{\fontsize{36 pt}{46.800000000000004 pt}\selectfont}}, color={rgb,1:red,0.0;green,0.0;blue,0.0}, draw opacity={1.0}, rotate={0.0}}, ymajorgrids={true}, ymin={-4}, ymax={45}, yticklabels={{$0$,$10$,$20$,$30$,$40$}}, ytick={{0.0,10.0,20.0,30.0,40.0}}, ytick align={inside}, yticklabel style={font={{\fontsize{36 pt}{46.800000000000004 pt}\selectfont}}, color={rgb,1:red,0.0;green,0.0;blue,0.0}, draw opacity={1.0}, rotate={0.0}}, y grid style={color={rgb,1:red,0.0;green,0.0;blue,0.0}, draw opacity={0.1}, line width={0.5}, solid}, axis y line*={left}, y axis line style={color={rgb,1:red,0.0;green,0.0;blue,0.0}, draw opacity={1.0}, line width={1}, solid}, colorbar={false}]
        \addplot[color={rgb,1:red,1.0;green,0.0;blue,0.0}, name path={c6be0337-0e8a-4e40-a85a-ccf66b9c561c}, area legend, fill={rgb,1:red,1.0;green,0.0;blue,0.0}, fill opacity={0.5}, draw opacity={0.5}, line width={0}, solid]
            table[row sep={\\}]
            {
                \\
                -15.0  -4.0  \\
                -6.0  -4.0  \\
                -6.0  1.0  \\
                -15.0  1.0  \\
            }
            ;
        \addlegendentry {$s \notin \psi$}
        \addplot[color={rgb,1:red,0.2745;green,0.5098;blue,0.7059}, name path={39bdae33-cb73-4d95-badd-b4c7db1c926f}, draw opacity={0.5891466906434875}, line width={1}, solid, forget plot]
            table[row sep={\\}]
            {
                \\
                -14.22075960598886  45.0  \\
                -12.108940934762359  37.88962019700557  \\
                -10.6800578335476  31.83514972962439  \\
                -9.483460396633848  26.49512081285059  \\
                -8.367753459681907  21.753390614533664  \\
                -7.579573609402271  17.56951388469271  \\
                -6.913117773508626  13.779727079991575  \\
                -7.1695703635462875  10.323168193237262  \\
                -6.849299272609858  6.7383830114641174  \\
                -5.517639580432713  3.3137333751591886  \\
                -6.1252244040854  0.5549135849428319  \\
                -6.1252244040854  0.5549135849428319  \\
            }
            ;
        \addplot[color={rgb,1:red,0.2745;green,0.5098;blue,0.7059}, name path={3f13f05a-1872-4367-913f-01c6aba3b852}, draw opacity={0.6666877295685241}, line width={1}, solid, forget plot]
            table[row sep={\\}]
            {
                \\
                -14.22075960598886  45.0  \\
                -12.31775582395494  37.88962019700557  \\
                -10.944250357931221  31.7307422850281  \\
                -9.284320275179937  26.258617106062488  \\
                -8.595383658562922  21.61645696847252  \\
                -7.665255072190615  17.318765139191058  \\
                -6.851455907089174  13.48613760309575  \\
                -5.582768640572134  10.060409649551163  \\
                -4.123385939524088  7.269025329265096  \\
                -4.42963345189216  5.207332359503052  \\
                -3.3712736771240186  2.9925156335569723  \\
                0.37739663839011417  1.306878794994963  \\
                -2.3391769476586903  1.49557711419002  \\
                -6.315573353244547  0.3259886403606749  \\
                -6.315573353244547  0.3259886403606749  \\
            }
            ;
        \addplot[color={rgb,1:red,0.2745;green,0.5098;blue,0.7059}, name path={83f4a3bb-67e8-415b-baa6-9e965290159e}, draw opacity={0.8575689314627534}, line width={1}, solid, forget plot]
            table[row sep={\\}]
            {
                \\
                -14.22075960598886  45.0  \\
                -12.162641888484359  37.88962019700557  \\
                -10.589927195807093  31.80829925276339  \\
                -9.186846233040683  26.513335654859844  \\
                -8.42131255581683  21.919912538339503  \\
                -7.767426091096679  17.70925626043109  \\
                -7.076195267154989  13.82554321488275  \\
                -6.993123882037646  10.287445581305255  \\
                -6.503836766906222  6.790883640286431  \\
                -5.1136113073215865  3.53896525683332  \\
                -6.185192477534791  0.9821596031725268  \\
                -6.185192477534791  0.9821596031725268  \\
            }
            ;
        \addplot[color={rgb,1:red,0.2745;green,0.5098;blue,0.7059}, name path={9bb3f821-df48-4c1a-8b57-a769e575dc2e}, draw opacity={1.0}, line width={1}, solid, forget plot]
            table[row sep={\\}]
            {
                \\
                -14.22075960598886  45.0  \\
                -12.143566912040114  37.88962019700557  \\
                -10.430442407406272  31.817836740985513  \\
                -9.085431227765124  26.602615537282375  \\
                -8.611258463141887  22.05989992339981  \\
                -7.842427547655474  17.754270691828868  \\
                -7.038909385479556  13.83305691800113  \\
                -6.416730401328431  10.313602225261352  \\
                -4.900150350888496  7.105237024597137  \\
                -4.625806161606518  4.655161849152888  \\
                -2.6231722148804018  2.342258768349629  \\
                -0.019893290707860167  1.030672660909428  \\
                -3.300718330904628  1.020726015555498  \\
                -6.34888767448731  -0.629633149896816  \\
                -6.34888767448731  -0.629633149896816  \\
            }
            ;
        \addplot[color={rgb,1:red,0.2745;green,0.5098;blue,0.7059}, name path={a55260ed-a4fd-40c5-bf7b-e2c6831e1fed}, draw opacity={0.0}, line width={1}, solid, forget plot]
            table[row sep={\\}]
            {
                \\
                -14.22075960598886  45.0  \\
                -12.142041629180312  37.88962019700557  \\
                -10.358022286311716  31.818599382415414  \\
                -9.17214773143007  26.639588239259556  \\
                -8.619351720881554  22.053514373544523  \\
                -7.946038312024107  17.743838513103746  \\
                -7.509445157530639  13.770819357091693  \\
                -7.28613428640268  10.016096778326373  \\
                -6.4646614334848245  6.373029635125033  \\
                -4.850708751546027  3.140698918382621  \\
                -6.174919770103951  0.7153445426096074  \\
                -6.174919770103951  0.7153445426096074  \\
            }
            ;
    \end{axis}
    \end{tikzpicture}}
    \caption{Failure trajectories for the lunar lander in altitude $y$ and vertical velocity $\dot{y}$ with HMC and baseline methods. Greater opacity reflects a higher likelihood failure.}
    \label{fig:lander-failures}
\end{figure}

\ph{Partially Observable Lunar Lander}
Failure trajectories for the lunar lander are shown in \cref{fig:lander-failures}. In the most likely failure mode, there is relatively small noise during the initial segment of the descent. A small amount of noise is added to the $AGL$ measurement in the negative direction, making the lander believe that it is slightly closer to the ground than it really is. Next, a larger amount of noise is added to $AGL$ in the positive direction, preventing the lander from increasing thrust in a critical moment. This highlights the ability of the proposed approach to find failures in non-Markov systems. The model-based approach is able to find diverse-likely failures by computing gradients through sequences of observations to find belief states that lead to failure.

\section{Conclusion}\label{sec:conclusion}
Estimating the distribution over failures is key to the safe and confident deployment of autonomous systems in safety-critical domains. This work frames the estimation of the distribution over failure trajectories in sequential systems as Bayesian inference. We take a model-based approach, where system gradients are used in HMC to sample disturbance trajectories that lead to failure.  Experiments against baseline approaches demonstrate that the proposed approach outperforms baselines in efficiently finding diverse, likely failure modes. The current approach is limited to systems compatible with AD, and is best suited when gradients are relatively inexpensive to compute. Future work will investigate using surrogate models \citep{qin2022statistical} or stochastic gradient approximations \citep{chen2014stochastic} for systems not compatible with AD, and approximate methods like variational inference \citep{kucukelbir2017automatic} when gradients are expensive.

\acks{This material is based upon work supported in part by the National Science Foundation Graduate Research Fellowship under Grant No. DGE-2146755. Any opinions, findings, and conclusions or recommendations expressed in this material are those of the authors and do not necessarily reflect the views of the National Science Foundation. This research was also supported in part by funding from Motional, Inc. This paper solely reflects the opinions and conclusions of its authors and not Motional or any other Motional entity.}

\bibliography{references}

\end{document}